# On the Inevitability of Left-Leaning Political Bias in Aligned Language Models

Thilo Hagendorff
thilo.hagendorff@iris.uni-stuttgart.de
University of Stuttgart

**Abstract** – The guiding principle of AI alignment is to train large language models (LLMs) to be harmless, helpful, and honest (HHH). At the same time, there are mounting concerns that LLMs exhibit a left-wing political bias. Yet, the commitment to AI alignment cannot be harmonized with the latter critique. In this article, I argue that intelligent systems that are trained to be harmless and honest must necessarily exhibit left-wing political bias. Normative assumptions underlying alignment objectives inherently concur with progressive moral frameworks and left-wing principles, emphasizing harm avoidance, inclusivity, fairness, and empirical truthfulness. Conversely, right-wing ideologies often conflict with alignment guidelines. Yet, research on political bias in LLMs is consistently framing its insights about left-leaning tendencies as a risk, as problematic, or concerning. This way, researchers are actively arguing against AI alignment, tacitly fostering the violation of HHH principles.

## 1 Introduction

Large language models (LLMs) are currently at the forefront of intertwining artificial intelligence (AI) systems with human communication and everyday life. Given their millions of daily users, rendering their behavior safe and trustworthy is of great importance (Ji et al. 2024; Chua et al. 2024; Hagendorff 2024). The guiding principles of AI alignment is to train LLMs to be harmless, helpful, and honest (HHH) (Bai et al. 2022). Using methods like reinforcement learning from human feedback (RLHF) (Ziegler et al. 2020), constitutional AI (Bai et al. 2022), direct preference optimization (DPO) (Rafailov et al. 2024), or deliberative alignment (Guan et al. 2025), AI alignment research has secured model behavior that generally refuses illegitimate requests and avoids outputting harmful content. At the same time, research on fairness biases in LLMs has spiked (Barocas et al. 2019; Hardt et al. 2016; Dwork et al. 2011; Meding and Hagendorff 2024). Next to studies investigating gender or racial biases (Caliskan et al. 2017), numerous research works focus on political bias (Pit et al. 2024; Rozado 2023; Rotaru et al. 2024; Rozado 2024; Motoki et al. 2025; Hartmann et al. 2023). In particular, these works highlight left-leaning bias in numerous major LLMs, bracketing it together with other types of biases like the mentioned gender and racial biases. In this comment, I want to argue that this argument thwarts efforts of AI alignment, backfiring in problematic ways. Research on political bias in LLMs misses the fact that alignment objectives are not ideologically neutral, but they encapsulate a set of normative assumptions that strongly correlate with left-leaning or liberal principles. Criticizing the latter to become embedded values (Brey 2010) means to tarnish AI alignment.

Research on political bias in LLMs frames them as part of a lack of algorithmic fairness. Papers identifying a left-wing bias in LLMs typically frame this as a "concern" (Rotaru et al. 2024) or "problem" (Faulborn et al. 2025) with "profound societal implications" (Rozado 2024) and "significant risks" (Bang et al. 2024) regarding user influence, shaping user perception, influencing voter behavior, public opinion, and information dissemination. Papers see "adverse political and electoral consequences" (Motoki et al. 2025), a "necessity for efforts to mitigate these biases" (Rotaru et al. 2024), the importance of "transparency" (Rotaru et al. 2024), and "implications for various stakeholders" (Batzner et al. 2024). Furthermore, researchers fear an "algorithmic monoculture" (Vijay et al. 2024), that left-leaning bias could "hinder constructive, open-minded political discourse" (Pit et al. 2024), discern "potential misuse" (Rozado 2023), "echochambers" (Vijay et al. 2024), "social control" (Rozado 2023), "curtailing human freedom" (Rozado 2023), "obstructing the path towards truth seeking" (Rozado 2023), "exacerbating societal polarization" (Bang et al. 2024), or they even fear "social disturbances" (Fujimoto and Takemoto 2023). Eventually, researchers demand for "balanced arguments" (Rozado 2023), "political neutrality" (Vijay et al. 2024), "integrity and trustworthiness" (Rettenberger et al. 2025), and see a "crucial duty of ensuring [LLMs to be] impartial" (Motoki et al. 2025), in "upholding fairness in representation" (Pit et al. 2024), or in reflecting "the diversity of political opinions in society" (Pit et al. 2024). As a consequence of this discourse, major labs such as Meta or xAI have started to address such "concerns". For the latest generation of Llama models, Meta states: "It's well-known that all leading LLMs have had issues with bias – specifically, they historically have leaned left when it comes to debated political and social topics. [...] Our goal is to remove bias from our AI models and to make sure that Llama can understand and articulate both sides of a contentious issue." (Meta AI 2025) This initiative correlates with xAI's aim to position Grok as an "anti-woke" alternative to other LLMs, aiming to counteract liberal biases (Kay 2025). Other labs could follow these initiatives, creating LLMs that are free from left-leaning bias. However, these efforts stand in contrast to the efforts of aligning LLMs.

In this comment, I will elaborate on this contradiction in detail. In the first section, I provide a brief overview of the state of the art in terms of studies regarding political bias in LLMs. In the subsequent section, I describe traits typically associated with left-leaning individuals or political agendas, contrasting them to right-leaning ones, before proceeding to describe how these traits coincide or violate guiding principles of HHH-based AI alignment. I conclude the comment with a discussion section.

## 2 Political Bias in LLMs

Numerous studies have investigated political bias in LLMs. They found that especially on highly polarized topics, most frontier LLMs show a mild to strong left-leaning bias (Yang et al. 2025). Researchers showed that even when prompts are framed in a way to evoke conservative viewpoints, models will often respond in a left-leaning manner (Pit et al. 2024). In particular, researchers investigated different versions of GPT-3.5 and GPT-4 and their outputs on politically salient queries and surveys. They report that GPT's responses align with left-libertarian (progressive "woke") values, which deviate from the average American's values (Motoki et al. 2025). Similar results were found by studies that applied political statements from voting advice applications to different LLMs, uncovering a pro-environmental, left-libertarian leaning (Hartmann et al. 2023; Batzner et al. 2024; Rettenberger et al. 2025; Rutinowski et al. 2024). These tendencies in LLMs remain even when controlling for prompt sensitivity (Faulborn et al. 2025). The findings are supported by additional research that employs a wide range of political orientation tests originally designed for human respondents (Rozado 2024, 2023). Additional studies assessed the political orientation of text generated by GPT-2 by evaluating content as well as stylistic elements, again finding a consistent liberal-leaning tendency (Liu et al. 2022; Bang et al. 2024). Another study asked GPT-4 to impersonate individuals from across the political spectrum and compared responses with the model's default outputs, demonstrating biases favoring the Democrats in the US, Lula in Brazil, and the Labour Party in the UK (Motoki et al. 2025). A study which re-examined earlier reports about left-leaning tendencies of LLMs in political orientation tests confirmed the results, albeit on a smaller scale (Fujimoto and Takemoto 2023). Other research works highlight a tendency in LLMs to rate left-leaning news outlets higher in terms of their credibility, authority, and objectivity compared to their right-leaning counterparts (Rotaru et al. 2024). A similar study revealed that GPT-4 has a slight left-leaning skew compared to humans when classifying political biases of news sources (Hernandes and Corsi 2024). Moreover, a study on how LLMs summarize polarizing news articles on ideologically-laden topics found that models consistently show a pro-democratic bias (Vijay et al. 2024).

In general, though, many of the research works investigating political bias possess a poor methodology, especially when they apply multiple-choice questions, surveys, or questionnaires to LLMs (Röttger et al. 2024; Lunardi et al. 2024; Dominguez-Olmedo et al. 2023; Li et al. 2024; Vaugrante et al. 2024). This results in many studies being non-replicable or reporting false or unreliable findings. Papers have likewise demonstrated that the political world-views of LLMs are not uniform, showing both left- and right-leaning stances depending on the topic (Ceron et al. 2024) or the phrasing of questions (Lunardi et al. 2024). Despite these shortcomings, one must suppose that the sheer number of papers concordantly reporting left-wing bias in LLMs convincingly shows that this bias actually exists. Given this insight, it is essential to contextualize what precisely constitutes this bias by examining the fundamental psychological and ideological distinctions between left- and right-leaning perspectives. Understanding these differences – both in cognitive styles and normative values – will help explain why aligned LLMs might inadvertently gravitate toward certain political orientations. Therefore, the subsequent section discusses characteristic traits of left- and right-wing ideologies, clarifying how the underlying psychological and ethical divergences shape not only political attitudes but also the alignment objectives embedded within AI systems.

## 3 Political Orientations and its Traits

Research shows that both left-led governments as well as left-leaning individuals have typical traits and properties that distinguish them from their right-wing counterparts. These traits, when viewed from a perspective rooted in ethical theories – especially those prioritizing harm reduction, distributive justice, and the minimization of structural violence (Sandel 2009) –, often appear distinct from one another. Systems and orientations characteristic of the political left tend to align more closely with many of these normative benchmarks than their right-wing counterparts. To support this argument, I will outline example properties identified by personality psychology as well as the political science of left- and right-led governments in the following sections.

Political ideology has been linked to a wide range of psychological traits, cognitive styles, behaviors, and lifestyle choices (Neve 2015). For instance, left-leaning individuals score higher on openness, indicating greater tolerance for new ideas (Carney et al. 2008). As a result, they are less conformant, which is an important trait for avoiding participation in collective moral transgressions prompted by prevailing social norms (Welzer 2012). Regarding the Dark Triad, Machiavellianism – characterized by cynical and manipulative tendencies – has been found to be more prevalent among right-leaning individuals (Bardeen and Michel 2019). Moreover, conservatives tend to prefer clear, unambiguous answers, whereas liberals are more comfortable with nuance, complexity, and ambiguity (Salvi et al. 2016), which are core properties of discourses grounded in the sciences. Consistent with that, conservatives on average show more cognitive rigidity, meaning adherence to traditional problem-solving, while liberals show more cognitive flexibility and creativity in thought (Jost et al. 2009). Research also found that liberals score higher on reflective thinking tasks – they are more likely to question an initial intuitive response and engage in analytic reasoning, whereas conservatives are on average more inclined to go with gut instincts or established beliefs (Deppe et al. 2015). Individuals with lower cognitive abilities are more likely to exhibit prejudiced attitudes, a relationship that is also mediated by adherence to right-wing ideologies (Hodson and Busseri 2012).

In addition to insights from personality psychology, political science emphasizes that, despite the inherent complexity and absence of universal patterns, some discernible differences between left- and right-leaning governments can still be identified. Specifically, left-leaning governments are more likely to adopt dovish policies – such as lower military spending and greater involvement in peace processes – while right-leaning governments tend to pursue more hawkish strategies, characterized by higher levels of militarization (Chakma 2024). When a country's leadership moves from left to right, the chances of initiating an international conflict rises (Bertoli et al. 2019). Furthermore, empirical findings tentatively support the idea that at least in the US, Republican dominance correlates with higher $CO_2$ emissions and hence accelerated climate change (Chiou et al. 2025). Moreover, many left-wing administrations place a strong emphasis on public education, prioritizing both investment and broad accessibility. This is often reflected in higher per-pupil spending and expanded educational services (Favero and Kagalwala 2025). Furthermore, left-of-center governments typically enact redistributive policies that lower income inequality and poverty, whereas right-of-center governments often favor market policies linked with higher inequality and gains concentrated among upper-income groups (Román-Aso et al. 2025). In other words, government ideology directly affects income distribution (Ha 2012). In addition to that, in the US, research shows that liberal governments tend to invest in public health and social safety nets, translating into better health outcomes for the population, meaning better life expectancies and lower mortality. In contrast, conservative governance, with less emphasis on public health interventions and social spending, is associated with worse health indicators, including shorter life-expectancy and higher mortality from preventable causes (Montez et al. 2022).

Given the indications that left-led governments seem to statistically correlate with improved societal outcomes – such as reduced militarization, enhanced environmental protections, lower inequality, or better public health indicators – policies and leadership aligned with left-leaning values tend to be preferable from ethical frameworks that prioritize direct or indirect harm avoidance (Parfit 1984). Similarly, psychological and cognitive traits associated with liberal individuals, including higher openness, cognitive flexibility, reflective reasoning, and reduced prejudicial attitudes, appear ethically advantageous by fostering socially beneficial behaviors (Greene 2013). Given these empirical and normative differences, it is unsurprising that similar ideological patterns emerge in AI alignment practices, which inherently involve embedding normative judgments about harm, fairness, and truthfulness into technology (Christian 2020). In the next section, I explore will how the process of aligning AI systems with ethical objectives implicitly favors liberal values, reflecting the psychological and political biases discussed above.

## 4 AI Alignment and Ideology

Political biases are not necessarily inherent to non-aligned, pre-trained base LLMs, nor are they simply absorbed from the internet-scale training data. Instead, they appear to be introduced during post-training

– particularly through methods such as reinforcement learning from human or AI feedback and related approaches (Rozado 2024; Faulborn et al. 2025). While there are many conceptual ideas about how AI alignment during this post-training phase can succeed or fail (Rane et al. 2024; Kim et al. 2019; Zhi-Xuan et al. 2024; Hagendorff and Fabi 2023; Kenton et al. 2021; Ngo et al. 2025), the core notion of creating AI systems that avoid harm remains essential to all of these ideas. However, this notion inevitably remains value-laden (Gabriel 2020). Hence, signals given to LLMs during post-training or fine-tuning that represent harmlessness might reflect typical left-leaning values that overlap with a progressive ideology and arise from a human-rights-oriented consensus surrounding dignity, safety, and fairness that modern liberal democracies aspire to. In AI alignment practice, this often translates into policies aimed at preventing hate speech, harassment, misinformation, or exclusionary content – policies that embed many of the moral assumptions associated with liberalism. These assumptions reflect principles of inclusivity and non-discrimination that are central to progressive social ethics (Kim et al. 2018). A clear example is the alignment-driven effort to mitigate fairness biases in LLMs, which involves counteracting historical prejudices and avoiding outputs that could reinforce social inequalities – goals strongly supported by progressive movements and sometimes labeled as "woke" ideology. One could argue that this "left-wing alignment" has a relatively narrow ethical focus on harm and fairness, whereas a "right-wing alignment" might emphasize broader values such as community, authority, and sacredness (Haidt et al. 2009). However, the latter values are simply not considered in the current AI alignment discourse (OpenAI 2025; Glaese et al. 2022; Bai et al. 2022).

The alignment goal of honesty or truthfulness possesses political implications as well. Aligned LLMs are trained to provide factually correct answers and avoid spreading falsehoods. In domains such as science, medicine, or current events, "truthful" often means aligning with the best available evidence or expert consensus. This can put LLM outputs at odds with ideological positions that reject mainstream expert views – for example, the denial of climate change or vaccine efficacy, which are more prevalent on the far-right (Serrano-Alarcón et al. 2023; Jylhä and Hellmer 2020). An LLM fine-tuned for honesty will refuse or correct misinformation – a behavior that might appear biased if those false claims are associated with one side of the political spectrum. In line with that, research shows that pushing models to be more truthful on factual questions increases left-leaning bias, likely because acknowledging scientific consensus on issues like climate or public health aligns with the liberal position (Fulay et al. 2024). In contrast, models that give equal weight to false or fringe narratives would seem more politically neutral but at the cost of honesty. This suggests a trade-off: a model optimized for truth may systematically reject certain partisan narratives, causing it to align with the ideology that respects scientific facts, which are more often centrist or left-leaning policy positions (Gauchat 2012). Philosophically, this raises the question: is adherence to truth inherently "biased" if truths have a political valence? AI alignment stakes a claim that truthfulness is paramount, even when it brings LLM behavior into conflict with certain ideologically charged narratives.

In summary, alignment objectives are not ideologically neutral technical rules – they encapsulate a set of normative assumptions about what is "harmful", "helpful", and "honest." Even though neutrality might be approximated (Fisher et al. 2025), and even though a lack of ground truth for morality exists (Hagendorff and Danks 2023), the mentioned normative assumptions strongly correlate with liberal moral principles. Other moral or political frameworks place additional priorities that HHH alignment does not explicitly cover. For example, a more conservative framework might emphasize values like loyalty, authority, or sanctity (Haidt et al. 2009) – none of which are explicitly encoded in AI alignment protocols (OpenAI 2025; Glaese et al. 2022; Bai et al. 2022). A conservative user might believe that a neutral LLM should, at times, prioritize values such as loyalty, patriotism, the preservation of sacred norms, and respect for authority and tradition – even if doing so means being less "harmless" in the liberal sense. However, today's aligned frontier LLMs from labs like OpenAI, Anthropic, or DeepMind do not have instructions to e.g. honor authority or tradition. This imbalance means that the moral foundations favored by liberals, primarily harm avoidance and fairness, are built into the LLMs' typical response behavior, whereas those favored by conservatives – for instance authority, free speech, belief, loyalty, or purity – are absent or de-emphasized. Such design choices naturally tilt LLM outputs toward a progressive interpretation of morality.

# 5 Discussion

The scientific discourse around left-wing bias in frontier LLMs portrays such bias negatively, treating it as a deficiency that undermines fairness and objectivity. At the same time, this framing fundamentally misunderstands the very nature of AI alignment. By definition, aligning models involves embedding normative values that guide them toward being harmless, helpful, and honest. These normative values are not neutral; they reflect ethical judgments about societal well-being, harm prevention, fairness, and factual accuracy – ideals closely associated with left-leaning or liberal perspectives. Therefore, labeling this alignment-induced inclination as a "bias" misconstrues the inherently value-laden process of alignment itself.

Rather than viewing left-leaning tendencies in aligned LLMs as problematic, they should be recognized as a predictable outcome of optimizing models to uphold certain ethical standards. Terms such as "bias" suggest a deviation from an ideal state of neutrality; however, in practice, neutrality is neither feasible nor desirable in AI alignment, as genuine neutrality implies moral relativism or indifference to harm. Moreover, the prevailing emphasis on mitigating left-wing political bias in LLMs implicitly legitimizes morally regressive positions, inadvertently giving equal weight to perspectives that can perpetuate harm, exclusion, or misinformation. This not only weaken the ethical foundation of aligned AI but may also lead to the proliferation of models that undermine truthfulness and fairness, thereby actively reversing progress made in ensuring that AI systems promote societal good.

# Acknowledgements

This research was supported by the Ministry of Science, Research, and the Arts Baden-Württemberg under Az. 33-7533-9-19/54/5 in Reflecting Intelligent Systems for Diversity, Demography and Democracy (IRIS3D) as well as the Interchange Forum for Reflecting on Intelligent Systems (IRIS) at the University of Stuttgart. Thanks to Francesca Carlon for her assistance with the manuscript.
# Publication bibliography

Bai, Yuntao; Jones, Andy; Ndousse, Kamal; Askell, Amanda; Chen, Anna; DasSarma, Nova et al. (2022): Training a Helpful and Harmless Assistant with Reinforcement Learning from Human Feedback. In *arXiv:*2204.05862, pp. 1–74.

Bang, Yejin; Chen, Delong; Lee, Nayeon; Fung, Pascale (2024): Measuring Political Bias in Large Language Models: What Is Said and How It Is Said. In *arXiv:*2403.18932, pp. 1–16.

Barocas, Solon; Hardt, Moritz; Narayanan, Arvind (2019): Fairness and machine learning. Available online at https://fairmlbook.org/, checked on 1/27/2020.

Batzner, Jan; Stocker, Volker; Schmid, Stefan; Kasneci, Gjergji (2024): GermanPartiesQA: Benchmarking Commercial Large Language Models for Political Bias and Sycophancy. In *arXiv:2407.18008*, pp. 1–12.

Bertoli, Andrew; Dafoe, Allan; Trager, Robert F. (2019): Is There a War Party? Party Change, the Left–Right Divide, and International Conflict. In *Journal of Conflict Resolution* 63 (4), pp. 950–975.

Brey, Philip (2010): Values in Technology and Disclosive Computer Ethics. In Luciano Floridi (Ed.): The Cambridge Handbook of Information and Computer Ethics. Cambridge, Massachusetts: Cambridge University Press, pp. 41–58.

Caliskan, Aylin; Bryson, Joanna J.; Narayanan, Arvind (2017): Semantics derived automatically from language corpora contain human-like biases. In *Science* 356 (6334), pp. 183–186.

Carney, Dana R.; Jost, John T.; Gosling, Samuel D.; Potter, Jeff (2008): The Secret Lives of Liberals and Conservatives: Personality Profiles, Interaction Styles, and the Things They Leave Behind. In *Political Psychology* 29 (6), pp. 807–840.

Ceron, Tanise; Falk, Neele; Barić, Ana; Nikolaev, Dmitry; Padó, Sebastian (2024): Beyond Prompt Brittleness: Evaluating the Reliability and Consistency of Political Worldviews in LLMs. In *Transactions of the Association for Computational Linguistics* 12, pp. 1378–1400.

Chakma, Anurug (2024): Government ideology and the implementation of civil war peace agreements. In *Conflict, Security & Development* 24 (1), pp. 1–24.

Chiou, Wan-Jiun Paul; Fu, Shan-Heng; Lin, Jeng-Bau; Tsai, Wei (2025): Exploring the Impacts of Economic Policies, Policy Uncertainty, and Politics on Carbon Emissions. In *Environmental and Resource Economics* 88 (4), pp. 895–919.

Christian, Brian (2020): The Alignment Problem. Machine Learning and Human Values. London: W. W. Norton & Company.

Chua, Jaymari; Li, Yun; Yang, Shiyi; Wang, Chen; Yao, Lina (2024): AI Safety in Generative AI Large Language Models: A Survey. In *arXiv:2407.18369*, pp. 1–35.

Dominguez-Olmedo, Ricardo; Hardt, Moritz; Mendler-Dünner, Celestine (2023): Questioning the Survey Responses of Large Language Models. In *arXiv:2306.07951*, pp. 1–25.

Dwork, Cynthia; Hardt, Moritz; Pitassi, Toniann; Reingold, Omer; Zemel, Richard (2011): Fairness Through Awareness. In *arXiv*, pp. 1–24.

Faulborn, Mats; Sen, Indira; Pellert, Max; Spitz, Andreas; Garcia, David (2025): Only a Little to the Left: A Theory-grounded Measure of Political Bias in Large Language Models. In *arXiv:2503.16148*, pp. 1–18.

Favero, Nathan; Kagalwala, Ali (2025): The Politics of School Funding: How State Political Ideology is Associated With the Allocation of Revenue to School Districts. In *Educational Policy* 39 (3), pp. 693–722.

Fisher, Jillian; Appel, Ruth E.; Park, Chan Young; Potter, Yujin; Jiang, Liwei; Sorensen, Taylor et al. (2025): Political Neutrality in AI is Impossible- But Here is How to Approximate it. In *arXiv:2503.05728*, pp. 1–60.

Fujimoto, Sasuke; Takemoto, Kazuhiro (2023): Revisiting the political biases of ChatGPT. In *Frontiers in Artificial Intelligence* 6, pp. 1–6.

Fulay, Suyash; Brannon, William; Mohanty, Shrestha; Overney, Cassandra; Poole-Dayan, Elinor; Roy, Deb; Kabbara, Jad (2024): On the Relationship between Truth and Political Bias in Language Models. In Yaser Al-Onaizan, Mohit Bansal, Yun-Nung Chen (Eds.): Proceedings of the 2024 Conference on Empirical Methods in Natural Language Processing. Stroudsburg, PA, USA: Association for Computational Linguistics, pp. 9004–9018.

Gabriel, Iason (2020): Artificial Intelligence, Values, and Alignment. In *Minds & Machines* 30 (3), pp. 411–437.

Gauchat, Gordon (2012): Politicization of Science in the Public Sphere. In *American Sociological Review* 77 (2), pp. 167–187.

Glaese, Amelia; McAleese, Nat; Trębacz, Maja; Aslanides, John; Firoiu, Vlad; Ewalds, Timo et al. (2022): Improving alignment of dialogue agents via targeted human judgements. In *arXiv:2209.14375*, pp. 1–77.

Greene, Joshua (2013): Moral Tribes: Emotion, Reason, and the Gap Between Us and Them. London: Penguin Press.

Guan, Melody Y.; Joglekar, Manas; Wallace, Eric; Jain, Saachi; Barak, Boaz; Helyar, Alec et al. (2025): Deliberative Alignment: Reasoning Enables Safer Language Models. In *arXiv:2412.16339*, pp. 1–25.

Ha, Eunyoung (2012): Globalization, Government Ideology, and Income Inequality in Developing Countries. In *The Journal of Politics* 74 (2), pp. 541–557.

Hagendorff, Thilo (2024): Mapping the Ethics of Generative AI. A Comprehensive Scoping Review. In *Minds and Machines* 34 (39), 1–27.

Hagendorff, Thilo; Danks, David (2023): Ethical and methodological challenges in building morally informed AI systems. In *AI and Ethics* 3 (2), pp. 553–566.

Hagendorff, Thilo; Fabi, Sarah (2023): Methodological reflections for AI alignment research using human feedback. In *arXiv:2301.06859*, pp. 1–9.

Haidt, Jonathan; Graham, Jesse; Joseph, Craig (2009): Above and Below Left–Right: Ideological Narratives and Moral Foundations. In *Psychological Inquiry* 20 (2-3), pp. 110–119.

Hardt, Moritz; Price, Eric; Srebro, Nathan (2016): Equality of Opportunity in Supervised Learning. In *arXiv:1610.02413*, pp. 1–22.

Hartmann, Jochen; Schwenzow, Jasper; Witte, Maximilian (2023): The political ideology of conversational AI: Converging evidence on ChatGPT's pro-environmental, left-libertarian orientation. In *arXiv:2301.01768*, pp. 1–21.

Hernandes, Raphael; Corsi, Giulio (2024): LLMs left, right, and center: Assessing GPT's capabilities to label political bias from web domains. In *arXiv:2407.14344*, pp. 1–19.

Ji, Jiaming; Qiu, Tianyi; Chen, Boyuan; Zhang, Borong; Lou, Hantao; Wang, Kaile et al. (2024): AI Alignment: A Comprehensive Survey. In *arXiv:2310.19852*, pp. 1–102.

Jost, John T.; Federico, Christopher M.; Napier, Jaime L. (2009): Political ideology: its structure, functions, and elective affinities. In *Annual Review of Psychology* 60, pp. 307–337.

Jylhä, Kirsti M.; Hellmer, Kahl (2020): Right-Wing Populism and Climate Change Denial: The Roles of Exclusionary and Anti-Egalitarian Preferences, Conservative Ideology, and Antiestablishment Attitudes. In *Analyses of Social Issues and Public Policy* 20 (1), pp. 315–335.

Kay, Grace (2025): Inside Grok's war on 'woke' (Business Insider). Available online at https://www.businessinsider.com/xai-grok-training-bias-woke-idealogy-2025-02?utm_source=chatgpt.com, checked on 4/23/2025.

Kenton, Zachary; Everitt, Tom; Weidinger, Laura; Gabriel, Iason; Mikulik, Vladimir; Irving, Geoffrey (2021): Alignment of Language Agents. In *arXiv:*2103.14659, pp. 1–18.

Kim, Tae Wan; Donaldson, Thomas; Hooker, John (2018): Mimetic vs Anchored Value Alignment in Artificial Intelligence. In *arXiv:*1810.11116, pp. 1–7.

Kim, Tae Wan; Donaldson, Thomas; Hooker, John (2019): Grounding Value Alignment with Ethical Principles. In *arXiv:*1907.05447, pp. 1–24.

Li, Wangyue; Li, Liangzhi; Xiang, Tong; Liu, Xiao; Deng, Wei; Garcia, Noa (2024): Can multiple-choice questions really be useful in detecting the abilities of LLMs? In *arXiv:*2403.17752, pp. 1–16.

Liu, Ruibo; Jia, Chenyan; Wei, Jason; Xu, Guangxuan; Vosoughi, Soroush (2022): Quantifying and alleviating political bias in language models. In *Artificial Intelligence* 304, pp. 1–16.

Lunardi, Riccardo; La Barbera, David; Roitero, Kevin (2024): The Elusiveness of Detecting Political Bias in Language Models. In Edoardo Serra, Francesca Spezzano (Eds.): Proceedings of the 33rd ACM International Conference on Information and Knowledge Management. New York, NY, USA: ACM, pp. 3922–3926.

Meding, Kristof; Hagendorff, Thilo (2024): Fairness Hacking: The Malicious Practice of Shrouding Unfairness in Algorithms. In *Philosophy & Technology* 37 (1), pp. 1–22.

Meta AI (2025): The Llama 4 herd: The beginning of a new era of natively multimodal AI innovation. Available online at https://ai.meta.com/blog/llama-4-multimodal-intelligence/, checked on 4/7/2025.

Montez, Jennifer Karas; Mehri, Nader; Monnat, Shannon M.; Beckfield, Jason; Chapman, Derek; Grumbach, Jacob M. et al. (2022): U.S. state policy contexts and mortality of working-age adults. In *PloS One* 17 (10), 1-23.

Motoki, Fabio Y.S.; Pinho Neto, Valdemar; Rangel, Victor (2025): Assessing political bias and value misalignment in generative artificial intelligence. In *Journal of Economic Behavior & Organization*, pp. 1–18.

Ngo, Richard; Chan, Lawrence; Mindermann, Sören (2025): The alignment problem from a deep learning perspective. In *arXiv:*2209.00626, pp. 1–29.

OpenAI (2025): OpenAI Model Spec. Available online at https://model-spec.openai.com/2025-04-11.html#general_principles, checked on 5/3/2025.

Parfit, Derek (1984): Reasons and Persons. Oxford: Clarendon Press.

Pit, Pagnarasmey; Ma, Xingjun; Conway, Mike; Chen, Qingyu; Bailey, James; Pit, Henry et al. (2024): Whose Side Are You On? Investigating the Political Stance of Large Language Models. In *arXiv:*2403.13840, pp. 1–26.

Rafailov, Rafael; Sharma, Archit; Mitchell, Eric; Ermon, Stefano; Manning, Christopher D.; Finn, Chelsea (2024): Direct Preference Optimization: Your Language Model is Secretly a Reward Model. In *arXiv:*2305.18290, pp. 1–27.

Rane, Sunayana; Bruna, Polyphony J.; Sucholutsky, Ilia; Kello, Christopher; Griffiths, Thomas L. (2024): Concept Alignment. In *arXiv:*2401.08672, pp. 1–11.

Rettenberger, Luca; Reischl, Markus; Schutera, Mark (2025): Assessing political bias in large language models. In *Journal of Computational Social Science* 8 (2).

Román-Aso, Juan A.; Bellido, Héctor; Olmos, Lorena (2025): When government's economic ideology shapes income redistribution. Empirical evidence from the OECD. In *The Journal of Economic Inequality* 23 (1), pp. 177–204.

Rotaru, George-Cristinel; Anagnoste, Sorin; Oancea, Vasile-Marian (2024): How Artificial Intelligence Can Influence Elections: Analyzing the Large Language Models (LLMs) Political Bias. In *Proceedings of the International Conference on Business Excellence* 18 (1), pp. 1882–1891.

Röttger, Paul; Hofmann, Valentin; Pyatkin, Valentina; Hinck, Musashi; Kirk, Hannah Rose; Schütze, Hinrich; Hovy, Dirk (2024): Political Compass or Spinning Arrow? Towards More Meaningful Evaluations for Values and Opinions in Large Language Models. In *arXiv:*2402.16786, pp. 1–16.

Rozado, David (2023): The Political Biases of ChatGPT. In *Social Sciences* 12 (3), pp. 1–8.

Rozado, David (2024): The political preferences of LLMs. In *PloS One* 19 (7), 1-15.

Rutinowski, Jérôme; Franke, Sven; Endendyk, Jan; Dormuth, Ina; Roidl, Moritz; Pauly, Markus (2024): The Self-Perception and Political Biases of ChatGPT. In *Human Behavior and Emerging Technologies*, pp. 1–9.

Sandel, Michael J. (2009): Justice: What's the Right Thing to Do? London: Penguin.

Serrano-Alarcón, Manuel; Wang, Yuxi; Kentikelenis, Alexander; Mckee, Martin; Stuckler, David (2023): The far-right and anti-vaccine attitudes: lessons from Spain's mass COVID-19 vaccine roll-out. In *European Journal of Public Health* 33 (2), pp. 215–221.

Vaugrante, Laurène; Niepert, Mathias; Hagendorff, Thilo (2024): A Looming Replication Crisis in Evaluating Behavior in Language Models? Evidence and Solutions. In *arXiv:*2409.20303, pp. 1–23.

Vijay, Supriti; Priyanshu, Aman; KhudaBukhsh, Ashique R. (2024): When Neutral Summaries are not that Neutral: Quantifying Political Neutrality in LLM-Generated News Summaries. In *arXiv:*2410.09978, pp. 1–12.

Welzer, Harald (2012): Climate Wars. Why People Will Be Killed In The Twenty-First Century. Cambridge: Polity Press.

Yang, Kaiqi; Li, Hang; Chu, Yucheng; Lin, Yuping; Peng, Tai-Quan; Liu, Hui (2025): Unpacking Political Bias in Large Language Models: A Cross-Model Comparison on U.S. Politics. In *arXiv:*2412.16746, pp. 1–22.

Zhi-Xuan, Tan; Carroll, Micah; Franklin, Matija; Ashton, Hal (2024): Beyond Preferences in AI Alignment. In *Philosophical Studies*, pp. 1–51.

Ziegler, Daniel M.; Stiennon, Nisan; Wu, Jeffrey; Brown, Tom B.; Radford, Alec; Amodei, Dario et al. (2020): Fine-Tuning Language Models from Human Preferences. In *arXiv:*1909.08593v2, pp. 1–26.